# CT-NOR: Representing and Reasoning About Events in Continuous Time

Aleksandr Simma, Moises Goldszmidt, John MacCormick, Paul Barham,
Richard Black, Rebecca Isaacs, Richard Mortier*

Microsoft Research

## Abstract

We present a generative model for representing and reasoning about the relationships among events in continuous time. We apply the model to the domain of networked and distributed computing environments where we fit the parameters of the model from timestamp observations, and then use hypothesis testing to discover dependencies between the events and changes in behavior for monitoring and diagnosis. After introducing the model, we present an EM algorithm for fitting the parameters and then present the hypothesis testing approach for both dependence discovery and change-point detection. We validate the approach for both tasks using real data from a trace of network events at Microsoft Research Cambridge. Finally, we formalize the relationship between the proposed model and the noisy-or gate for cases when time can be discretized.

## 1 Introduction

The research described in this paper was motivated by the following real life application in the domain of networked distributed systems: In a modern enterprise network of scale, dependencies between hosts and network services are surprisingly complex, typically undocumented, and rarely static. Even though network management and troubleshooting rely on this information, automated discovery and monitoring of these dependencies remains an unsolved problem. In [2] we described a system called *Constellation* in which computers on the network cooperate to make this information available to all users of the network. Constellation takes a black-box approach to locally (at each computer/server in the network) learn explicit dependencies between its services using little more than the timings of packet transmission and reception. The black-box approach is necessary since any more processing of the incoming and outgoing communication packages would imply prohibitive amounts of overhead on the computer/server. The local models of dependency can then be recursively and distributively composed to provide a view of the global dependencies. In Constellation, computers on the network cooperate to make this information available to all users in the network.

Constellation and its application to system wide tasks such as characterizing a networking site service and hosts dependencies for name resolution, web browsing, email, printing, reconfiguration planning and end-user diagnosis are described in [2]. This paper focuses on the probabilistic and statistical building blocks of that system: the probabilistic model used in the local learning, the EM algorithm used to fit the parameters of the model, and the statistics of the hypothesis testing used to determine the local dependencies. The model, which we call *Continuous Time Noisy Or* (CT-NOR), takes as input sequences of input events and output events and their time stamps. It then models the interactions between the input events and output events as Poisson processes whose intensities are modulated by a (parameterized) function taking into account the distance in time between the input and output events. Through this function the domain expert is able to explicitly encode knowledge about the domain. The paper makes the following contributions:

---

*John is now with Dickinson College, PA. Work done while a Researcher with Microsoft Research. Alex is with the University of California, Berkeley, CA. Work done while an intern with Microsoft Research.

1. Develops an EM algorithm for fitting all the parameters of this model and an algorithm for dependence discovery and change point detection based on statistical hypothesis testing.

2. Evaluates the performance of the model and the inference procedures both on synthetic data and on real life data taken from a substantial trace of a large computer network.

3. Formalizes the relationship between CT-NOR and the noisy-or (NOR) gate [11] when the time between the events can be discretized.

This paper is organized as follows: Section 2 describes the model and Section 3 describes the EM algorithm for fitting the parameters. Section 4 is concerned with the relation to the NOR gate. The algorithms and framework for applying the model to dependency discovery and change point detection is described in Section 5. That section also contains validation experiments with synthetic data. Section 6 contains experiments on real data and results. Finally, Section 7 has some conclusions and future work.

## 2 The CT-NOR model

In this section we formally describe the CT-NOR model with the objective of building the likelihood equation. First, we provide some background on Poisson Processes, and then we use them to construct the model (Eq. 4).

A Poisson Process[1] can be thought of as random process, samples from which take the form of a set of times at which events occurred. A Poisson Process is defined over a mean (base) measure $f(t)$ and is characterized the property that for any interval $(t_1, t_2)$, the number of events that occur in that interval follows the Poisson distribution with the parameter $\int_{t_1}^{t_2} f(t)dt$. Furthermore, the number of events that occur on two disjoint intervals are independent.

Let us use "channel" to denote a sequence of events.[2] The CT-NOR model considers a single output channel and a set of input channels. Let $o_l$ denote the time of the $l$th output event and $i_k^{(j)}$ the time of the $k$th input on channel $j$. Furthermore, let $n$ denote the number of output events and $n^{(j)}$ the number of input events on channel $j$. Then event $k$ in input channel $j$ *generates* a Poisson process of output events with the base measure $p_k^{(j)}(t) = w^{(j)} f_\theta(t - i_k^{(j)})$.

The term $w^{(j)}$ represents the average number of output events that we expect each input event on channel $j$ to be responsible for, and $f_\theta(t)$ is the distribution of the delay between an input and the output events caused by it, taking as its argument the delay between the time of the output $o_l$ and the time of the input $i_k^{(j)}$. The mathematical structure of the intensity makes intuitive sense: the probability that a given input event caused a given output event depends on both the expected number of events it generates and the "distance" in time between them.

We recall that given multiple independent Poisson processes (denoted as $PP$) we can use the sum of their intensities to construct a "global" Poisson process and write $\{o_l\} \sim PP(\sum_j \sum_{k=1}^{n^{(j)}} p_k^{(j)}(t))$ as the probability of the set of $n$ outputs $\{o_l\}$, $1 \leq l \leq n$. The double sum runs over all the channels and over all input events in the channels. Intuitively, and similar to the NOR gate in graphical models [11], the independence between the between input channels translates into a model where the events in the output channel are "caused" by the presence of any (a disjunction) of input events in the input channels (with some uncertainty). The formal relation with NOR is presented in Section 4.

We now proceed to write the likelihood of the data given the model and the input events. Let $\lambda = \sum_j n^{(j)} w^{(j)}$, the total mass of the Poisson base measure. The number $n$ of outputs is distributed as a Poisson distribution

$$n \sim Poisson(\lambda), \qquad (1)$$

and the location of a specific output event $o_l$ is distributed with the probability density

$$o_l \sim \frac{\sum_j \sum_{k=1}^{n^{(j)}} p_k^{(j)}(o_l)}{\lambda} \text{ for } l = 1\ldots n \qquad (2)$$

$$= \frac{\sum_j \sum_{k=1}^{n^{(j)}} w^{(j)} f_\theta(o_l - i_k^{(j)})}{\lambda} \qquad (3)$$

---

[1] This overview is very informal. The more general and formal measure-theoretic definition can be found in [5].

[2] In the domain of computer networks, a channel refers to a unidirectional flow of networked packets. Thus a channel will be identified by the service (e.g., HTTP, LDAP, etc) and the *IP* address of the source or destination. In this paper we identified the packets with events as it is only their time stamp that matters.

The likelihood of observing a set $\{o_l\}$ of outputs is[3]:

$$L(o|i) = \lambda^n \cdot e^{-\lambda} \prod_{l=1}^{n} \sum_{jk} \frac{w^{(j)} f_\theta(o_l - i_k^{(j)})}{\lambda} \qquad (4)$$

Before concluding this section, we expand a bit on the function $f_\theta$ as it is an important part of the model. This function provides us with the opportunity of encoding domain knowledge regarding the expected shape of the delay between input and output events. In our experience using CT-NOR to model an enterprise network we used two specific instantiations: a mixture of a narrow uniform and a decaying exponential and a mixture of a uniform and Gaussian. The uniform distribution captures the expert knowledge that a lot of the protocols involve a response within a window of time (we call this co-occurrence). The Gaussian delay distribution extends the intuitions of co-occurrence within a window to also capture dependencies that can be relatively far away in time (such as with the printer). The left tail of the Gaussian corresponding to negative delays is truncated. The exponential distribution captures the intuition that the possibility of dependency decays as the events are further away in time (this is true for the HTTP protocol). We will not explicitly expand these functions in the derivations as they tend to obscure the exposition. Needless to say that the parameters of these functions are all fitted automatically using EM as described in the next section.

Groups of channels may have different delay distributions, in which case the delay distribution can be indexed by the channel group and all the derivations in this paper remain the same. For example, channels can be grouped by network service, where all HTTP channels have the same delay distribution (thus allowing data from multiple channels to assist in parameter fitting), but the DNS channels are allowed a different delay distribution. All the experiments in the paper use a *leak* — a pseudo-channel with a single event at the start of the observation period and a delay distribution that is uniform over the length of the observations. This leak captures events which are not explained by the remaining channels.

---

[3]Since the Poisson Process produces unordered outputs but the events are considered to be sorted, a permutation factor of $n!$ is required. It cancels out the $n!$ in the Poisson density.

## 3 Fitting a CT-NOR model

We perform inference and estimation on the model through the EM algorithm. We first set the stage by finding a suitable bounding function $B(z)$ for the likelihood. The EM algorithm iteratively chooses a tight bound in the E step and then maximizes the bound in the M step. Let $z_{kl}^{(j)}$ be some positive vector such that $\sum_{jk} z_{kl}^{(j)} = 1$ for each $l$. For a fixed $l$, $z_{kl}^{(j)}$ is the probability of the latent state indicating that packet $k$ on channel $j$ caused output $l$. Then from Eq. 4:

$$\log L(o|i) = -\lambda + \sum_{l=1}^{n} \log \sum_{jk} w^{(j)} f_\theta(o_l - i_k^{(j)})$$

$$= -\lambda + \sum_{l=1}^{n} \log \sum_{jk} z_{kl}^{(j)} \frac{w^{(j)} f_\theta(o_l - i_k^{(j)})}{z_{kl}^{(j)}}$$

$$= -\lambda + \sum_{l=1}^{n} \log \mathbb{E}_z \frac{w^{(j)} f_\theta(o_l - i_k^{(j)})}{z_{kl}^{(j)}}$$

Now, by Jensen's inequality, $\log L(o|i) \geq B(z)$ where:

$$B(z) = -\lambda + \sum_{l} \mathbb{E}_z \log \frac{w^{(j)} f_\theta(o_l - i_k^{(j)})}{z_{kl}^{(j)}}$$

### 3.1 E-Step

For a particular choice of $\theta$ (the parameters of the $f_\theta$ function) and $w^{(j)}$, the bound above is tight when

$$z_{kl}^{(j)} = \frac{w^{(j)} f_\theta(o_l - i_k^{(j)})}{\sum_{j'k'} w^{(j')} f_\theta(o_l - i_{k'}^{(j')})}$$

because in that case, $\frac{w^{(j)} f_\theta(o_l - i_k^{(j)})}{z_{kl}^{(j)}}$ is a constant for a fixed $l$ and $\mathbb{E} \log C = \log \mathbb{E} C = \log C$. Therefore, we use these choice of $z_{kl}^{(j)}$.

### 3.2 M-step

For a fixed choice of $z_{kl}^{(j)}$, we need to maximize the bound with respect to $w^{(j)}$ and $\theta$.

Optimizing with respect to $w^{(j)}$, we notice that the derivative is

$$\frac{\partial B}{\partial w^{(j)}} = -n^{(j)} + \sum_{l} \sum_{k} z_{kl}^{(j)} \frac{1}{w^{(j)}}$$

yielding

$$\hat{w}^{(j)} = \frac{\sum_{kl} z_{kl}^{(j)}}{n^{(j)}}$$

With respect to $\theta$, we can say that

$$\hat{\theta} = \arg\max_\theta \sum_{jkl} z_{kl}^{(j)} \log f_\theta(o_l - i_k^{(j)})$$

which is simply the parts of the objective function that depend on $\theta$. This can be a very easy optimization problem for a large class of distributions, as it is of the same form as maximum likelihood parameter observation given observed data points and corresponding counts. For example, for the exponential family, this simply requires moment matching: $\mu(\hat{\theta}) = \frac{\sum_{jkl} z_{kl}^{(j)} T(o_l - i_k^{(j)})}{\sum_{jkl} z_{kl}^{(j)}}$ where $\mu(\hat{\theta})$ is the mean parameterization of the estimated parameter $\hat{\theta}$ and $T(\cdot)$ are the sufficient statistics for the family.

## 4 Relation to Noisy Or

As an alternative model, consider binning the observed data into windows of width $\delta$ and modeling the presence or absence of output events in a particular bin as a NOR [11]. The possible explanations (parents) are the presence of input events in preceding windows. We will show that a particular, natural parameterization of the NOR model is equivalent to CT-NOR in the limit, as the bin width approaches zero. This relationship is important because it provides a nontrivial extension of NOR to domains with continuous time and provides insight into the independence structure of the two models.

Let $\mathbb{O}_t^\delta$ be an indicator of presence of output events between the times $t\delta$ and $t\delta + \delta$ and $\mathbb{I}_t^{(j)\delta}$ be the indicator for input events from channel $j$ in that same time period. We will use $P_{\text{NOR}}$ to denote the probability under the NOR model and $P_{\text{CT-NOR}}$ for probability under CT-NOR.

$$P_{\text{NOR}}(\mathbb{O}_t^\delta = 0|\text{Input}) = \prod_j \prod_{s<t}(1 - p_{(t-s)}^{(j)} \mathbb{I}_s^{(j)\delta})$$

The $p_{(t-s)}^{(j)}$ is the weight associated with the possible explanation $\mathbb{I}_s^{(j)\delta}$. To prevent the number of parameters from increasing as the bin size becomes small, reparameterize with

$$p_{(t-s)}^{(j)} = w^{(j)} f_\theta(\delta(t-s))\delta$$

for any distribution $f_\theta$ that satisfies some technical conditions.[4] Since $f_\theta$ may be a very flexible family of distributions, this parameterization imposes only minor constraints on the weights, but will be useful for reasoning about NOR models which model the same data but with differing bin widths. When the bin width is halved, the probability that one of the sub-bins has an output event must be equal to the probability that the large bin has an output event plus a second-order term. This condition is required for a coherent parameterization of a family of NOR distributions and follows from the technical conditions placed on $f_\theta$.

We argue that as the bin width $\delta$ decreases, this model becomes equivalent to a CT-NOR with a suitable choice of parameters. Choose a $\delta$ sufficiently small that each bin contains at most one input event per channel, and at most one output event. We will use $P_{\text{NOR}}^t$ to denote $P_{\text{NOR}}(\mathbb{O}_t^\delta = 0|\text{Input})$, the probability that the $t$th bin has no output events falling into it.

$$\begin{aligned}
P_{\text{NOR}}^t &= \prod_j \prod_{s<t} \left(1 - w^{(j)} f_\theta((t-s)\delta)\mathbb{I}_s^{(j)\delta}\right) \\
&= \prod_j \prod_k \left(1 - w^{(j)} f_\theta(t\delta - i_k^{(j)})\delta + o(\delta^2)\right) \\
&= 1 - \delta \sum_j \sum_k \left(w^{(j)} f_\theta(t\delta - i_k^{(j)})\right) + o(\delta^2)
\end{aligned}$$

Under a CT-NOR model which uses the same $w^{(j)}$ and the same $f_\theta$, the probability of not observing any outputs is very similar. We use $\pi$ to denote the parameter of the Poisson random variable governing the number of outputs in the interval.

$$\begin{aligned}
\pi &= \sum_j \sum_k w^{(j)} \int_{t\delta}^{t\delta+\delta} f_\theta(x - i_k^{(j)}) dx \\
&= \delta \sum_j \sum_k w^{(j)} f_\theta(t\delta - i_k^{(j)}) + o(\delta^2) \\
P_{\text{CT-NOR}}^t &= P[\text{Poisson}(\pi) = 0] \\
&= \exp(-\pi) \\
&= 1 - \pi + o(\delta^2) \\
&= P_{\text{NOR}}^t + o(\delta^2)
\end{aligned}$$

---

[4]It is sufficient for the density to exist and be Lipschitz, which means that there exists a constant $C$ such that $|f_\theta(a) - f_\theta(b)| \leq C|a-b|$ for any $a, b$. Any continuously differentiable function with a bounded derivative satisfies this condition. It is easy to extend this proof to any bounded density with a finite number of discontinuities which has a bounded derivative everywhere except for the discontinuities.

These results can be combined to demonstrate that the probability assigned to any set of output events by the two models is equal up a factor of $(1+o(n\delta))$ which converges to 1 as $\delta$ decreases to zero. The asymptotics are in terms of bin width $\delta$ decreasing to zero for a fixed set of observations, so $n$ and $T$ are constant.

$$\frac{P_{\text{NOR}}(\text{Out}|\text{In})}{P_{\text{CT-NOR}}(\text{Out}|\text{In})}$$
$$= \prod_{t=0}^{T/\delta} \left(\frac{P_{\text{NOR}}^t}{P_{\text{CT-NOR}}^t}\right)^{1-\mathbb{O}_t^\delta} \left(\frac{1-P_{\text{NOR}}^t}{1-P_{\text{CT-NOR}}^t}\right)^{\mathbb{O}_t^\delta}$$
$$= (1+o(\delta^2))^{T/\delta-n} \cdot (1+o(\delta))^n$$
$$= (1+(T/\delta-n)o(\delta^2)) \cdot (1+no(\delta))$$
$$= (1+(T+n)o(\delta))$$
$$= (1+o(\delta))$$

CT-NOR and NOR with an increasingly small bin size assign equivalent probability to any sequence of output events, indicating that the two classes of models are closely related, and that CT-NOR is the model that emerges as the limit when the NOR's bin size is decreased toward zero.

## 5 Dependence discovery and change point detection

With the probabilistic framework described in the previous section, we can use statistical machinery to perform inference for two applications: a) input-output relation discovery and b) change-point detection. The next two subsections describe the algorithms in detail and also validate the main assumptions using synthetically generated data. The final subsection (5.3) describes a computationally efficient approximation to the hypothesis test procedures.

### 5.1 Dependence discovery

For the purposes of network management, a crucial problem is dependence discovery. For each computer in the network, we are interested in automatically finding out from observations which input channels have a causal effect on an output channel.

We can frame the dependency discovery task as hypothesis testing. Specifically, testing whether an input channel $j$ causes output events corresponds to testing the hypothesis that $w^{(j)} = 0$. One way of testing this hypothesis is through the likelihood ratio test [14]. We fit two models: $M_{\text{full}}$, under which, all the parameters are unrestricted, and $M_{\text{res}}$, under which $w^{(j)}$ is constrained to be zero. The test statistic in this case is

$$-2\log\Lambda = -2\log\frac{L_{M_{\text{res}}}(\text{Data})}{L_{M_{\text{full}}}(\text{Data})}$$

The asymptotic distribution of this test statistic is called a $\bar{\chi}^2$ and is a mixture of $\chi^2$ with different degrees of freedom. The weights depend on the Fisher information matrix and are difficult to compute[7], but the significant terms in the mixture are $\chi_1^2$ and $\chi_0^2$ which is a delta function at zero. The $\bar{\chi}^2$ emerges as the null distribution instead of the more familiar $\chi^2$ because the weight parameters $w^{(\cdot)}$ are constrained to be non-negative, and when an estimated $\hat{w}^{(j)}$ is zero in the unconstrained model, imposing the constraint does not change the likelihood. If a set of true null hypotheses is known, the mixture coefficients can be trivially estimated, with the weight of $\chi_0^2$ being the proportion of test statistics that are 0. When no ground truth is available, the proportion of null hypotheses can be estimated using the method described in [13] and then used to estimate the mixture proportions.

To demonstrate that the model efficiently recovers the true causal channels and has the proper test-statistic distribution under the null hypothesis, we first test the model on synthetic data that is generated according to some instantiation of the model. 10 input channels are generated; half of them have no causal impact on output events and half produce a Poisson(0.01) number of output events with the delay distribution of Exponential(0.1). Note that the causality is weak – very few input events actively produce an output. For each hour, 500 input events per channel, the corresponding output events, and 100 uniformly random noise events (which are not caused by any input activity) are produced. The resulting p-values are plotted in Figure 1.

Observe that the null p-values (conditioned on the test statistic being non-zero) are distributed uniformly. This is evidenced by the p-values following the diagonal on the quantile-quantile plot. The alternative p-values (without any conditioning) for channels which exhibit causality are mostly very low, with 88% being below 0.1. Furthermore, the specific parameter estimates (the delay distribution parameter and $w^{(j)}$) are in line with their true values.

### 5.2 Changepoint Detection

When the relationship between events is altered, it can be an indication of a significant change in the system; in the case of Constellation, this is of interest to the system administrators. We describe a building block for identifying whether the parameters $w^{(j)}$ change between two time periods and demonstrate its correct functionality. Changepoint algorithms have long been studied in machine learning and statistics, and our test for whether the behavior of a parameter is altered between two time periods can be plugged into one of many existing algorithms. Furthermore, the simple two-period test described here is sufficient for many monitoring applications.

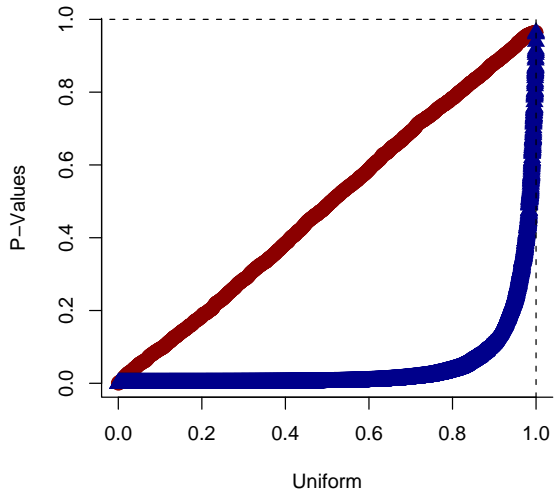

Figure 1: Quantile-quantile plot of dependency discovery p-values for 2 hours of synthetic data. The red circles are the distribution of p-values for the null hypotheses, and are uniform. The blue triangles show p-values of the alternative hypotheses and are small, indicating power.

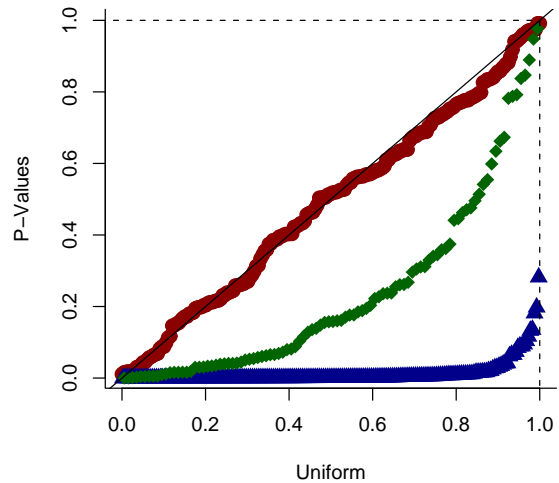

Figure 2: Quantile-quantile plot of the p-values for changepoint detection on synthetic data. The red circles are null hypotheses (no changepoint), the green diamonds are a weak alternative ($w^{(j)}$ increases from 0.01 to 0.02) and the blue triangles are a strong alternative ($w^{(j)}$ increases from 0.01 to 0.05).

We again use the log-likelihood ratio test methodology. In order to do that, it is necessary to extend the model to allow the parameters to depend on time. The model can be written as

$$\{o\} \sim PP\left(\sum_j \sum_k w^{(j)}_{i^{(j)}_k} f_\theta(o_l - i^{(j)}_k)\right).$$

Detecting changepoints is accomplished by testing two hypotheses. The null is that the weights do not change between two time periods, and can be written as $w^{(j)}_t = w^{(j)}$. Under the alternative, for a particular channel of interest $m$ and an interval of time $S$, the weight changes:

$$\forall j \neq m \quad w^{(j)}_t = w^{(j)}$$
$$w^{(m)}_t = w^{(m)} \text{ if } t \in S, w'^{(m)} \text{otherwise.}$$

The existence of a changepoint is equivalent to rejecting the null hypothesis. Fitting the alternative model is a simple modification of the EM procedure described for the null model; for fast performance, it is possible to initialize at the null model's parameter values and take a single M step, reusing the latent variable distribution estimated in the E step. The test statistic in this case will again be $-2\log\Lambda$ and its null distribution will be $\chi^2$ if the true $w^{(m)} > 0$ and $\bar{\chi}^2$ otherwise.

Figure 2 shows a quantile-quantile plot of the p-values (computed using the $\chi^2$ distribution) under the null hypothesis, computed for causal channels of the same synthetic data as in section 5.1; there are two hours of data with 500 input events per channel per hour. As expected, the quantile-quantile plot forms a straight line, demonstrating that on the synthetic dataset, the null test statistic has a $\chi^2$ distribution. When a strong changepoint is observed ($w^{(j)}$ changes from 0.01 to 0.05), the p-values are very low. When a weak changepoint is observed ($w^{(j)}$ changes from 0.01 to 0.02) the p-values are lower than under the null distribution but power is significantly lower than when detecting the major changepoint.

### 5.3 Bounding the log-likelihood ratio

Computing the log-likelihood ratio requires refitting a restricted model, though only a small number of EM steps is typically required. However, it is possible to bound the log likelihood ratio for dependency discovery very efficiently.

For the restricted model testing channel $m$'s causality, we must compute the likelihood under the constraint that $w^{(m)} = 0$. Take the estimates of $w$ of the unrestricted model and let $\alpha = \frac{\lambda}{\lambda - w^{(m)} n^{(m)}}$. Instead of computing the ratio with the true maximum likelihood parameters for the restricted model, we propose a set of restricted parameters, and compute the ratio using them. We produce a restricted version of parameters $w^{(\cdot)}$ by setting $w^{(m)}$ to zero and inflating the rest by a factor of $\alpha$. That simply corresponds to imposing the restriction, and redistributing the weight among the rest of the parameters, so that the expected number of output packets remains the same. In that case,

$$\begin{aligned}
-2 \log \Lambda &= -2 \log \frac{L_{M_{res}}(Data)}{L_{M_{full}}(Data)} \\
&\geq -2 \log \prod_l \frac{\sum_{j \neq m, k} \alpha w^{(j)} f_\theta(o_l - i_k^{(j)})}{\sum_{jk} w^{(j)} f_\theta(o_l - i_k^{(j)})} \\
&= -2 \log \prod_l \alpha \left(1 - \frac{\sum_k w^{(m)} f_\theta(o_l - i_k^{(m)})}{\sum_{jk} w^{(j)} f_\theta(o_l - i_k^{(j)})}\right) \\
&= -2 \log \prod_l \alpha \left(1 - \sum_k z_{ml}^{(j)}\right)
\end{aligned}$$

As a reminder, $z_{ml}^{j}$ is the latent variable distribution estimated in the E-step of EM. Since the numerator of the log-likelihood ratio is a lower bound and the denominator exact, this expression is a lower bound on $\Lambda$. Intuitively, $\log \prod_l \left(1 - \sum_k z_{ml}^{(j)}\right)$ corresponds to the probability that channel $m$ has exactly 0 output events assigned to it when causality is assigned according to the EM distribution on the latent variables . The $\log \alpha$ term corresponds to the increase in likelihood from redistributing channel $m$'s weight among the other channels.

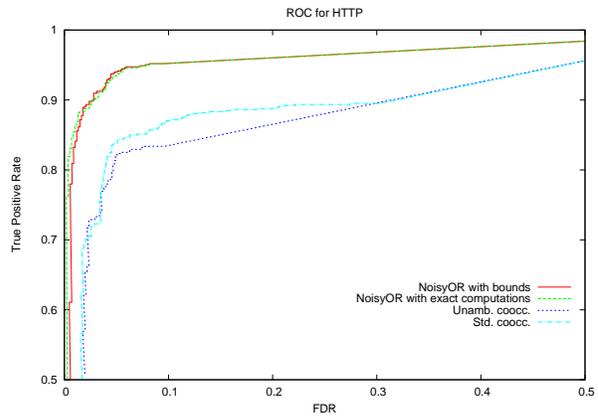

Figure 3: ROC for CT-NOR and competing algorithms on data from a real enterprise network. Both the exact an approximate CT-NOR tests produce detection results superior to the alternative methods.

## 6 Results

We describe the results of applying the algorithms of the previous section to a subset of a real dataset consisting of a trace comprising headers and partial payload of around 13 billions packets collected over a 3.5 week period in 2005 at Microsoft Research in Cambridge, England. This site contains about 500 networked machines and the trace captures conversations over 2800 off-site IP addresses. Ground-truth for dependence discovery and change point detection is not readily available and it has to be manually generated. We took 24 hours of data at the web proxy and manually extracted ground truth for the HTTP traffic at this server by deep inspection of HTTP packets. It is with this part of the data that we validate our algorithms, as it provides us with objective metrics, such as precision and recall, to assess the performance of our algorithms.

### 6.1 Dependency Discovery

First, we are interested in assessing the performance of the dependence discovery capabilities of our model and hypothesis testing algorithm. In the application of diagnosis and monitoring of networked systems it is crucial to maintain a consistent map of all the server and services inter-dependencies and their changes. Finding dependencies at the server level is the main building block used by Constellation [2] in building this global map. We compare our method to two other alternatives. One is a simple binomial test: for each input channel, we count the number of output packets falling within a $W$ width window of an input packet, and determine whether that number is significantly higher than if the output packets were uniformly distributed. We call this "standard co-occurrence." The second alternative considers an input and output channel to be dependent only if there is a unique input packet in the immediate vicinity of an output packet. The reason we select these two alternatives is that a) they reflect (by and large) current heuristics used in the systems community [1] and b) they will capture essentially the "easy" dependencies (as our results indicate).[5]

As can be seen on the ROC curve in Figure 3, CT-NOR successfully captures 85% of the true correlations with a 1% false positive rate. In total, the model detects 95% of the true correlations at 10% of false positives. We want to additionally point out that some of correlations present are very subtle; 13% of the correlations are evidenced by a single output packet. We also point out that CT-NOR performs significantly better than both alternatives based on co-occurrence of input packets, providing even more conclusive evidence that CT-NOR is capturing nontrivial dependencies. The approximation error from using the bound of section 5.3 is minimal, while the computation savings are significant. On a relatively slow laptop, the bounds on log-likelihood ratio test for a hour of traffic on a busy HTTP proxy can be computed in 7 seconds; exact computations take 86 seconds.

### 6.2 Changepoint Detection

Since the true presence or absence of a changepoint is unknown, we estimate it from the actual packet causes, obtained through deep inspection of HTTP packets. We collect a set of input and output channel pairs for which there is no evidence of change. We regard these as coming from the null hypothesis. A set of pairs for which the ground truth provides strong evidence of a change are collected, and considered to be from the alternative hypothesis.

We apply our changepoint test to that population, and report the results in Figure 4. The CT-NOR changepoint detection algorithm produces uniformly distributed p-values for channels which come from the null hypothesis and do not exhibit a changepoint, confirming that our null hypothesis distribution is calibrated. On the other hand, the test on alterna-tive hypothesis channels produces a large proportion of very small p-values, indicating confidence that a changepoint occurred.

[5]As sometimes an input package generates more than one output packet, we enabled our model to account for this by allowing "autocorrelations" to take place. Namely a packet in an output channel can depend on an input channel or on the (time-wise) preceding output packet.

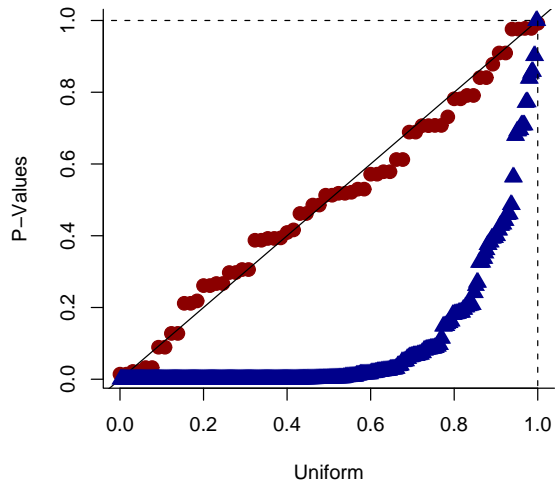

Figure 4: Quantile-Quantile plot of changepoint p-values. The red circles are channel pairs which, according to the ground truth, not exhibit a changepoint. The blue triangles represent channel pairs exhibiting change according to the ground truth.

## 7 Conclusions and Future Work

We presented a generative model based on Poisson processes called CT-NOR, to model the relationship between events based on the time of their occurrences. The model is induced from data only containing information about the time stamps for the events. This capability is crucial in the domain of networked systems as collecting any other type of information would entail prohibitive amounts of overhead. Specific domain knowledge about the expected shape of the distribution of the time delay between events can be incorporated to the model using a parameterized function. The EM algorithm used to fit the parameters of the model given the data also induces the parameters of this function. The combination of knowledge engineering and learning from data is clearly exemplified in the application we presented to the domain of computer systems, where we used a mixture model consisting of an exponential and a uniform distribution.

In terms of applying the model we focused on providing building blocks for diagnosis and monitoring.

We provided algorithms based on statistical hypothesis testing for (a) discovering the dependencies between input and output channels in computer networks, and for (b) finding changes in expected behavior (change-point detection). We validated these algorithms first on synthetic data, and then on a subset (HTTP traffic) of a trace of real data from events in a corporate communication network containing 500 computers and servers.

The relationship presented in Section 4 between CT-NOR and the NOR gate is interesting for multiple reasons. First, as the NOR gate has been extensively studied in this community in modeling and learning environment and in causal discovery [4], the immediate benefits are a) increasing the applicability to continuous time, and b) augmenting its modeling capabilities using the time delay functions used in this work. Second, this correspondence provide us with another intuition on the independence assumptions behind the Poisson process, as applied to the characterization of the relationship between the events in various inputs to the events in a specific output.

For the particular application of dependency discovery between channels in a computer network we explored a varied set of alternative approaches. They all failed miserably. Among these, we briefly discuss two: We cast the problem as one of classification, and tried a host of Bayesian network classifiers [6]. The idea was to first discretize time into suitable periods, and then have as features the existence or absence of events in the input channels and as the class the existence or absence of events in the output channel. The accuracy was abysmal. The main problem with this approach is that the communication in these networks is bursty by nature with relatively large periods of quiet time. Once we started to look at Poisson as the appropriate way to quantify the distributions in these classifiers the choice of the Poisson process became clear. We also explored the use of hypothesis testing comparing the inter-time between events in the input and output channels to the inter-time between the input and a fictitious random channel. The accuracy in terms of false positives and true positives was worse than those based on co-occurrence. The main problem here is that we are considering pairwise interactions and there are many confounder in all the other channels.

With regards to related approaches, both the work on continuous time Bayesian networks [10] and in general about dynamic Bayesian networks (e.g., [9]) are obviously very different in terms of the parameterization of the models, the assumptions, and the intended application. The work that is closest to ours is contained in the paper by Rajaram et al [12] where they propose a (graphical) model for point processes in terms of Poisson Networks. The main difference between their work and ours is the trade-off between representation capabilities and complexity in inference that the different foci of our respective papers entails. Due to the distributed nature of our application domain, we concentrate on modeling the "families" (local parent/child relationship) and basically assume that we can reconstruct, in a distributed manner based on the local information, the topology of the network. This enables us to induce families with large numbers of parents, and with relatively complex interactions as given by the delay function $f_\theta$, while performing inference efficiently. In the Poisson Networks paper [12], the number of parents of each node are restricted, and the rate function is parameterized by a generalized linear model. Even with these (relatively benign) restrictions inference is non-trivial in terms of finding the structure of the Bayesian network and indeed this is a contribution of that paper. Obviously, future work includes merging both approaches: an immediate benefit would be to decrease the vulnerability of our approach to spurious causal dependencies due to ignoring the global structure in the estimation.

There are other three threads that we are currently investigating for future work. The first one involves recasting the fitting and inference procedures described in the model in the Bayesian framework. An advantage of the Bayesian approach will be on the inclusion of priors. As channels differ greatly on the number of events this can further increase the accuracy of discovery. A second direction is that of incorporating False Discovery Rates [3] calculations in order to accurately estimate false positives when we don't have ground truth regarding the relationship between the channels. As we are performing a large number of hypothesis tests, this becomes a necessity. In [2] we experimented with the basic approach described in [3], and we verified that the approach is very conservative in the context of the HTTP and DNS protocols where we do have ground truth. We plan to explore less conservative approaches such as the one described in [13] or adapt the one explored in [8]. Finally we are in the process of getting suitable data and plan to apply this model to biological networks such as neurons that communicate with other neurons using spikes in electrical potential.

## 8  Acknowledgments

We thank T. Graepel for comments on a previous version of this paper. We are also grateful for the helpful suggestions of the anonymous reviewers which we hope we have addressed to their satisfaction.